\documentclass{article} 
\usepackage{iclr2025_conference,times}

\usepackage{amsmath,amsfonts,bm}

\def\eqref#1{equation~\ref{#1}}

\def\1{\bm{1}}

\DeclareMathAlphabet{\mathsfit}{\encodingdefault}{\sfdefault}{m}{sl}
\SetMathAlphabet{\mathsfit}{bold}{\encodingdefault}{\sfdefault}{bx}{n}

\usepackage[utf8]{inputenc} 
\usepackage[T1]{fontenc}    
\usepackage{hyperref}       
\usepackage{url}            
\usepackage{booktabs}       
\usepackage{amsfonts}       
\usepackage{nicefrac}       
\usepackage{microtype}      
\usepackage{xcolor}         
\usepackage{amsmath}
\usepackage{graphicx}

\usepackage{pifont} 
\usepackage{multirow}

\def\eg{\emph{e.g.\ }}

\usepackage{cleveref}
\crefname{table}{table}{tables}
\Crefname{table}{Table}{Tables}
\crefname{figure}{figure}{figures}
\Crefname{figure}{Figure}{Figures}
\crefname{equation}{equation}{equations}
\Crefname{Equation}{Equation}{Equation}

\usepackage{tikz}

\title{Predicting 3D representations for Dynamic Scenes}

\iclrfinalcopy

\author{%
  Di Qi$^{1}$\thanks{Equal contribution.} \hfill Tong Yang$^{32*}$\thanks{Corresponding authors.} \hfill Beining Wang$^{3*}$\thanks{This work is done during Beining Wang's internship at StepFun.} \hfill Xiangyu Zhang$^{12 \dag}$ \hfill Wenqiang Zhang$^{3}$\\[2mm]
  $^1$StepFun~~~~$^2$MEGVII Technology~~~~$^3$Fudan University\\[1pt]
  {\small\texttt{
    \{qidi, wangbeining\}@stepfun.com
  }} \hfill {\small\texttt{\{yangtong, zhangxiangyu\}@megvii.com}}
}

\begin{document}

\maketitle

\begin{abstract}
We present a novel framework for dynamic radiance field prediction given monocular video streams. Unlike previous methods that primarily focus on predicting future frames, our method goes a step further by generating explicit 3D representations of the dynamic scene. The framework builds on two core designs. First, we adopt an ego-centric unbounded triplane to explicitly represent the dynamic physical world. Second, we develop a 4D-aware transformer to aggregate features from monocular videos to update the triplane. Coupling these two designs enables us to train the proposed model with large-scale monocular videos in a self-supervised manner. Our model achieves top results in dynamic radiance field prediction on NVIDIA dynamic scenes, demonstrating its strong performance on 4D physical world modeling. Besides, our model shows a superior generalizability to unseen scenarios. Notably, we find that our approach emerges capabilities for geometry and semantic learning.
\end{abstract}

\section{Introduction}
As humans, we can perceive physical world, plan our behaviors and interact with environments. All of these rely on our ability of 4D physical modeling, especially 3D future prediction. For example, when playing football, we predict the 3D motion of the ball and opponents to determine our next move. However, despite enormous achievements in computer vision, most studies focus on 2D or 3D tasks, with limited exploration into 3D future prediction. This gap hinders the development of critical fields, such as autonomous driving, robotics and augmented reality.

Having the ability of 3D future prediction, machine learning models should exhibit two characteristics: 1) generalized 3D representation; 2) 4D prior learning. While there are many 3D representation choices, such as Neural Radiance Fields (NeRFs) and 3D Gaussian Splatting, these representations are primarily designed for object- or scene-centric reconstruction. Using these representations to represent a large and complex object/scene leads to a heavy memory and computation cost. For effective 4D prior learning, models must acquire the ability to predict 3D structures at future times based on historical data. This requires exposure to large-scale datasets that contains 4D knowledge about the physical world. Unfortunately, such 3D/4D-like data is scarce, limiting the learning of 4D prior knowledge. 

To resolve above issues, we firstly propose an ego-centric unbounded triplane that enables object- and scene-agnostic modeling. Our 3D representation adopts the viewpoint of observer (camera) as the central reference point, ensuring independence from specific objects or scenes. Also, since the physical world is inherently unbounded, we incorporate a non-linear contraction mechanism from mip-NeRF 360~\citep{barron2022mip} to handle it. Furthermore, we leverage explicit triplanes as our 3D representation due to their compact and structured design. Secondly, to overcome the scarcity of 3D data, we utilize monocular videos as our training data, which inherently encode 3D information, as demonstrated by existing models like SORA~\citep{videoworldsimulators2024}. Training on these monocular videos allows our model to acquire 4D prior knowledge by predicting explicit 3D representations at a future time using sequences of video frames. 

Based on the above analysis, we introduce a framework for dynamic radiance field prediction, as shown in~\Cref{fig:overview}. Specifically, we initialize an unbounded ego-centric triplane for the target time $t=S+1$. Then, given a sequence of monocular video frames at time points $\{t\}_{t=1}^{S}$, we use a 4D-aware transformer to update the triplane features. The updated triplane represents the dynamic radiance filed at the target time $t=S+1$. To train the model, we render the target view from the updated triplane with volume rendering and optimize the difference between the rendered target view and its ground-truth counterpart. Additionally, to ensure better 3D consistency, we introduce a temporal-based 3D constraint, applying photometric loss to two target views rendered at different times from the same sequence of video frames.

We conduct extensive experiments to demonstrate the effectiveness of our method. Our method exhibits strong 3D future prediction performance on the NVIDIA Dynamic Scenes and generalizes well to the DAVIS dataset, which presents a larger domain shift from our training data. This confirms the capacity of our method to learn generalized 4D prior knowledge in a data-driven manner. Moreover, we find emergent capabilities of our model, including geometry and semantic learning. 

In summary, our approach predicts dynamic radiance fields, explicit 3D representations, at future times, enabling the ability of 4D physical world modeling. We validate our method by conducting extensive experiments on various settings, especially generalization on unseen scenarios. Moreover, our method can pave the way for “next-3D prediction” paradigm, which is a promising path for spatial intelligence. 

\section{Related Work}
\subsection{Radiance Field Rendering}
Radiance field rendering has recently achieved a remarkable progress. Neural Radiance Fields (NeRFs)~\citep{mildenhall2021nerf} leverage MLPs to represent scenes and employ volume rendering to produce high-quality images from novel viewpoints. The success of NeRF has inspired numerous subsequent works that address its limitations~\citep{barron2022mip, chan2022efficient, muller2022instant, yu2021plenoctrees, chen2022tensorf} and expand its applications~\citep{poole2022dreamfusion, wang2024prolificdreamer, hong2023lrm}. Mip-NeRF360~\citep{barron2022mip} demonstrates impressive view synthesis results on unbounded scenes, while EG3D~\citep{chan2022efficient} and InstantNGP~\citep{muller2022instant} use triplanes and hash grids, respectively, to accelerate computation. However, these methods inevitably face a trade-off between speed and quality. To overcome this challenge, 3D Gaussian Splatting~\citep{kerbl20233d} is proposed. This method represents scenes using 3D Gaussians, projects them into 2D through a rasterization mechanism and renders image similarly to NeRF. Thanks to its real-time speed and high-quality outputs, numerous follow-up works~\citep{yan2023multi, fu2023colmap, szymanowicz2023splatter, ling2023align} have rapidly emerged. ~\citep{yan2023multi} employs multi-scale 3D Gaussians to mitigate aliasing, while CF-3DGS~\citep{fu2023colmap} enables novel view synthesis without any SfM preprocessing by utilizing explicit point clouds and the continuity of input video stream. Existing neural rendering methods mainly focus on object- or scene-centric reconstruction, suffering from expensive cost for large-scale objects and scenes. In contrast, our approach adopts an ego-centric scene modeling, prioritizing the viewpoint of the observer (camera) to efficiently represent dynamic scenes. 

\subsection{Neural Fields for Dynamic Scenes}
Dynamic 3D scene modeling plays a pivotal role in various applications, ranging from AR/VR to autonomous driving~\citep{yang2023emernerf, zhou2023drivinggaussian, yang2023unisim}. Numerous works have been proposed to tackle this challenge~\citep{park2020deformable, pumarola2021d, wang2022fourier, li2022neural, fridovich2023k, cao2023hexplane}. For instance, D-NeRF~\citep{park2020deformable} and Nerfies~\citep{pumarola2021d} represent scenes by mapping observed points into a canonical scene representation using a volumetric deformation field. Similarly, DyNeRF~\citep{li2022neural} and K-Planes~\citep{fridovich2023k} compress dynamic scenes into implicit or explicit NeRFs, enabling scene rendering conditioned on position, view direction, and time. To overcome the reliance on multi-view data, some approaches~\citep{li2023dynibar, tian2023mononerf, Zhao2024PGDVS} focus on modeling dynamic scenes using monocular videos. For example, DynIBaR~\citep{li2023dynibar} synthesizes novel views by aggregating image features from nearby frames in a motion-aware manner. However, DynIBaR~\citep{li2023dynibar} is tailored to specific-scene optimization and struggles to generalize to unseen scenarios. While MonoNeRF~\citep{tian2023mononerf} and PGDVS~\citep{Zhao2024PGDVS} aim to enhance generalization, they still require scene-specific optimization or fine-tuning when applied to new scenes. Unlike previous works, which focus on reconstructing dynamic 3D scenes, our method introduces a novel approach for predicting dynamic 3D scenes at future times based on sequences of monocular video. 

\section{Method}
\begin{figure*}[t]
	\begin{center}
		\includegraphics[width=1.0\linewidth]{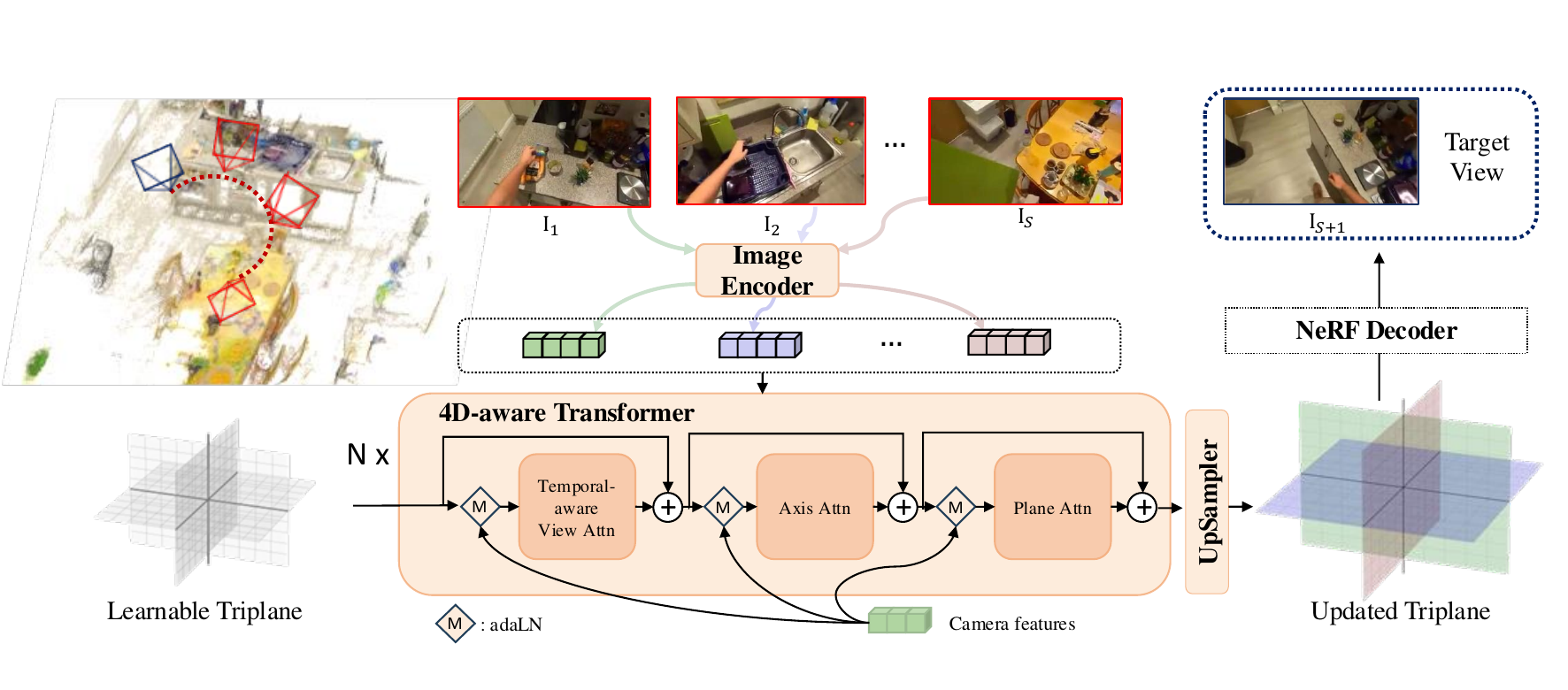}
	\end{center}
	\caption{\textbf{Overview of our method.} The process starts with an image encoder that extracts 2D image features from the the video frames, serving as a prior. These features are then processed by a 4D-aware transformer, which updates a learnable triplane representation. Next, an upsampler refines and enlarges the triplane. During training, a NeRF decoder generates target views via volumetric rendering and photometric loss is applied to optimize the model. \textit{The FFN module in the transformer is hidden in the figure.}}
	\label{fig:overview}
\end{figure*}

In this work, our method focuses on predicting 3D scene representations at future times from steaming monocular videos. Concretely, as illustrated in~\Cref{fig:overview}, we first employ an image encoder to extract features from the input monocular video. These features are then aggregated using a 4D-aware transformer to update an ego-centric triplane representation, Finally, the triplane is upsampled to produce the final 3D presentation at the target time. The model is optimized using photometric loss between the ground-truth views and target views rendered by a NeRF decoder. Additionally, we incorporate a temporal-based 3D constraint to enhance the consistency of the 3D scene representations. 

\subsection{Ego-centric Triplane Representation}
\label{sec:representation} 
Existing 3D representations are typically object-  or scene-centric. However, when modeling large objects or scenes, these representations impose significant memory and computational costs. To address this issue, we propose an ego-centric approach for constructing 3D representations, focusing on the perspective of the observer. Unlike object- or scene-centric representations, ego-centric representations are independent of specific objects and scenes, concentrating solely on the observer's view of the environment.

Since 3D scene prediction needs to generalize across diverse scenes, the predicted 3D representation should be both structured and fixed in size. To achieve this, we adopt a triplane as the foundation of the ego-centric 3D representation. The ego-centric triplane $\mathbf{T}$ consists of three axis-aligned feature planes $\mathbf{T}_{xy}, \mathbf{T}_{yz}$, and $\mathbf{T}_{xz}$. Each plane has dimensions $H \times W \times D$, where $H \times W$ represents the spatial dimensions and $D$ denotes the feature channels. To represent unbounded scenes with bounded representation, we transform the unbounded ego coordinates into bounded triplane coordinates. Concretely, we apply a non-linear contraction function from mip-NeRF 360~\citep{barron2022mip}, defined as:
\begin{equation}
\mathcal{C}(\mathrm{x})= \begin{cases}\mathrm{x} & \|\mathrm{x}\| \leq 1 \\ \left(2-\frac{1}{\|\mathrm{x}\|}\right)\left(\frac{\mathrm{x}}{\|\mathrm{x}\|}\right) & \|\mathrm{x}\|>1\end{cases}
\end{equation}
where $\mathrm{x} \in \mathbb{R}^3$ is 3D point in observer coordinates, $\mathcal{C}(\mathrm{x})$ is the point in triplane coordinates.

\subsection{Dynamic Radiance Field Predictor}

\subsubsection{Image Encoder}
Given a sequence of video frames $\{\mathbf{I}_{t}\}_{t=1}^{S}$, we employ a hybrid neural network to extract image features $\{\mathbf{F}_{t}\}_{t=1}^{S}$. For simplicity and efficiency, this hybrid network comprises a ResNet-like backbone and a self-attention layer. The ResNet-like backbone downsamples the spatial dimensions of the input images by a factor of 8, while the self-attention layer enhances the long-range contextual understanding of the image features. This enhancement  supports motion-aware feature aggregation in temporal-aware view-attention module~\ref{view-attn}. It is worth noting that we train the image encoder from scratch and observe that it learns semantic information through the 3D representation prediction task. This highlights our approach as an alternative method for self-supervised learning, see~\ref{sec:sl}. Furthermore, pre-trained vision foundation models, such as DINO~\citep{zhang2022dino} and MAE~\citep{he2022masked}, could also be used as the image encoder. However, we leave this exploration for future work. 

\subsubsection{4D-aware Transformer}
\begin{figure*}[t]
	\begin{center}
		\includegraphics[width=1.0\linewidth]{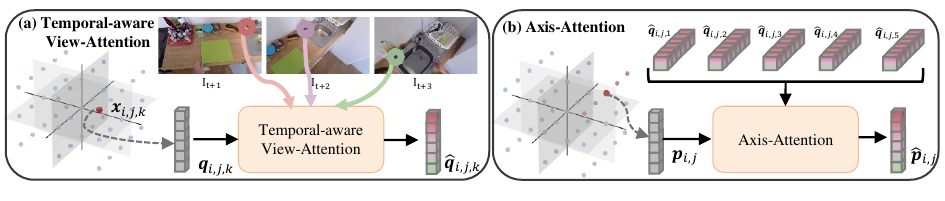}
		\vspace{-8mm}
	\end{center}
	\caption{\textbf{Temporal-aware View-Attention and Axis-Attention modules in Transformer.}
\textbf{(a) Temporal-aware View-Attention Module}: At the target time $t=S+1$, 3D virtual points are uniformly sampled within the triplane. For a given point $\mathbf{x}_{i,j,k}$, it is projected along the three axes onto the triplane features  $\mathbf{T}_{xy}, \mathbf{T}_{yz}$, and $\mathbf{T}_{xz}$ to obtain the corresponding 3D query feature $\mathbf{q}_{i,j,k}$. Simultaneously, $\mathbf{x}_{i,j,k}$ is mapped onto image feature maps to obtain epipolar features $\{\mathbf{f}_t\}_{t=1}^{S}$ from source video frames. The temporal-aware view-attention module then integrates these epipolar features across different time points $\{t\}_{t=1}^{S}$, producing an updated 3D query feature $\hat{\mathbf{q}}_{i,j,k}$.
\textbf{(b) Axis-Attention Module}: For a triplane feature $\mathbf{p}_{i,j}$ at pixel $(i, j)$ located in plane $\mathbf{T}_{xy}$, it is associates with point features along the z-axis $\{\hat{\mathbf{q}}_{i,j,k}\}_{k=1}^L$. The axis-attention module aggregates these point features to generate a refined triplane feature $\hat{\mathbf{p}}_{i, j}$.}
	\label{fig:attention module}
\end{figure*}
We initialize with a shared learnable ego-centric triplane. To update the triplane features, we introduce a 4D-aware transformer that aggregates information from monocular video features $\{\mathbf{F}_{t}\}_{t=1}^{S}$. The 4D-aware transformer consists of three key components: the temporal-aware view-attention module, the axis-attention module, and the plane-attention module.

\paragraph{Temporal-aware View-Attention Module}\label{view-attn}
The temporal-aware view-attention module is designed to aggregate the image features $\{\mathbf{F}_{t}\}_{t=1}^{S}$ from the source video frames in a motion-aware manner, as shown in~\Cref{fig:attention module}~(a). Specifically, we uniformly sample 3D virtual points in triplane space, denoted as $\{\mathbf{x}_{i,j,k}\}\begin{smallmatrix}M&N&L\\i=1&j=1&k=1\end{smallmatrix}$, where $i$, $j$ and $k$ represent x, y and z axes, $M$, $N$ and $L$ denote the number of points along each axis, respectively. For a virtual point $\mathbf{x}_{i, j,k}$, its feature $\mathbf{q}_{i,j,k}$ is computed by projecting it onto three planes of the triplane, querying the corresponding point features, and concatenating these features. To obtain its epipolar point feature $\mathbf{f}_t$ at time $t$, we project $\mathbf{x}_{i,j,k}$ onto the image feature at time $t$ using camera pose $\mathbf{P}_t$ and compute the feature via bilinear interpolation on the image feature grids. To derive the updated virtual point feature $\hat{\mathbf{q}}_{i,j,k}$, we use the virtual point feature $\mathbf{q}_{i,j,k}$ as a query and its corresponding epipolar point features $\{\mathbf{f}_t\}_{t=1}^{S}$ as keys and values. Additionally, we also consider the target time $t=S+1$ and other time points $\{t\}_{t=1}^{S}$ as time embeddings for the queries and keys to compute attention scores. Consequently, the updated virtual point feature $\hat{\mathbf{q}}_{i,j,k}$ is calculated as:
\begin{equation}
\hat{\mathbf{q}}_{i,j,k} = CrossAttn(\mathbf{q}_{i,j,k}, S+1, \{\mathbf{f}_t\}_{t=1}^{S}, \{t\}_{t=1}^{S})+\mathbf{q}_{i,j,k}
\end{equation}

Empirical results show that our temporal-aware view-attention module can implicitly distinguish between dynamic and static contents. This is validated by the temporal-aware similarities between the 3D virtual point feature $\mathbf{q}_{i,j,k}$ and its epipolar point features $\{\mathbf{f}_t\}_{t=1}^{S}$, see Appendix~\ref{sec: view-attention}. Since epipolar constraint is invalid in dynamic scenes, the virtual point features of dynamic content contain significant noise. To address this issue, we propose the axis-attention module and the plane-attention module to refine virtual point features.

\paragraph{Axis-Attention Module}
The cost of directly applying self-attention to refine virtual point features is substantial. Therefore, we introduce the axis-attention module to project new virtual point features onto the triplane. In the axis-attention module, we utilize triplane features as queries and their corresponding virtual point features as keys and values. For instance, as depicted in~\Cref{fig:attention module}~(b), we take the triplane feature $\mathbf{p}_{i,j}$ on $\mathbf{T}_{xy}$ as the query and the associated virtual point features $\{\hat{\mathbf{q}}_{i,j,k}\}_{k=1}^L$ along the $z$ axis as keys and values. Furthermore, we incorporate a position bias (3D position embedding) to each head when computing similarity. Finally, we compute the new triplane feature $\hat{\mathbf{p}}_{i, j}$ on plane $\mathbf{T}_{xy}$ using cross-attention, formulated as follows:
\begin{equation}
\hat{\mathbf{p}}_{i,j} = CrossAttn(\mathbf{p}_{i,j}, \{\hat{\mathbf{q}}_{i,j,k}\}_{k=1}^L)+\mathbf{p}_{i,j}
\end{equation}

\paragraph{Plane-Attention Module}
In the temporal-aware view-attention module, the aggregated features may not adequately capture dynamic contents, as the epipolar constraint becomes invalid in dynamic scenes. To address this limitation, we introduce a plane-attention module that leverages 3D-related information. Similar 
 to~\citep{cao2023large}, our plane-attention module incorporates both self- and cross-plane attention. The self-plane attention focuses on enhancing the semantic information of individual planes, while the cross-plane attention aims to establish generalized 3D prior knowledge across different planes. This allows us to utilize learned semantic information and 3D prior knowledge to correct feature errors arising from the temporal-aware view-attention module. 

Specifically, the goal of self-plane attention is to update each plane feature by aggregating features from the same plane. In cross-plane attention, each plane feature uses itself as query, while other planes serve as keys and values. Additionally, we introduce a position bias for each head when computing similarity. It is important to note that our ego-centric triplane is designed to represent unbounded scenes. In general, we need to embed the infinite position, but it is hard to implement explicitly. To circumvent this issue, we employ learnable position embeddings to represent position bias.

\paragraph{Camera Feature}
The training dataset comprises images captured with a wide range of focal lengths, leading to scale ambiguity. To mitigate this, we use camera intrinsic matrix as an inductive bias. We construct a camera feature  $\mathbf{c} \in \mathbb{R}^{16}$ for each target view by flattening its 4-by-4 camera intrinsic matrix. This feature $\mathbf{c}$ is then embedded by mapping it into a higher-dimensional space using a sinusoidal function $\gamma (\cdot)$ and projecting it into input dimension via a multi-layer perceptron (MLP). Inspired by DiT~\citep{peebles2023scalable}, we incorporate adaptive layer normalization (adaLN) within our feature attention block to effectively constrain the inputs of each attention block based on camera features.

\subsubsection{Upsampler}
To enhance the resolution of the triplane, we employ a trainable deconvolution layer to upscale the triplane embeddings $\mathbf{T}_{xy}, \mathbf{T}_{yz}$ and $\mathbf{T}_{xz}$ extracted from the transformer. After the upsampling process, we obtain a 3D representation at the target time.

\subsection{Training}

\paragraph{NeRF Decoder}
To optimize our model, we utilize the photometric loss between the rendered target views and ground-truth (GT) views. The rendered target views are obtained from the ego-centric triplane through volume rendering. Specifically, we employ NeRF as the decoder to predict both color (RGB) and density ($\sigma$) based on the 3D point features extracted from the triplane. Initially, we normalize the 3D positions using the contraction function described in Section \ref{sec:representation} and project them onto the three planes. We then concatenate the features from these planes to form the final feature vector, which is decoded into color (RGB) and density ($\sigma$) using a lightweight multi-layer perception (MLP). 

\paragraph{Temporal-based 3D Constraint}
To enhance the 3D consistency of the ego-centric triplane, we introduce a temporal-based 3D constraint during training. Specifically, given a sequence of source views $\{\mathbf{I}_t\}_{t=1}^{S}$, we predict two 3D representations at time $t=0$ and time $t=S+1$, respectively. Based on these two 3D representations, we render their corresponding target views and apply the photometric loss on these two rendered views. The assumption behind this constraint is that these two rendered views share dynamic scene overlap, which can be used to learn 3D geometry.

\paragraph{Training Strategy}
We adopt a cost-effective training strategy by starting with low-resolution input images and progressively increasing to higher resolutions. Initially, we pretrain our model on $128 \times 72$ images until convergence. Subsequently, we fine-tune it using $512 \times 288$ images, significantly reducing computational costs while still achieving superior quality compared to direct high-resolution training with equivalent computational resources.
\paragraph{Training Objective}
To mitigate the high cost of rendering full-resolution images for volume rendering, we randomly select $64 \times 64$ and $128 \times 128$ patches from target images with resolutions of $128 \times 72$ and $512 \times 288$, respectively, for loss supervision. We evaluate the visual accuracy of our renderings against the ground-truth (GT) images using three types of losses: L2 reconstruction loss $\mathcal{L}_\mathrm{recon}$, perceptual loss $\mathcal{L}_\mathrm{lpips}$, and structural similarity loss $\mathcal{L}_\mathrm{ssim}$. Additionally, to address the challenges posed by semi-transparent clouds, we apply a regularization term $\mathcal{L}_\mathrm{dist}$, inspired by the distortion loss in Mip-NeRF360~\citep{barron2022mip}. The overall training loss function is formulated as follows: 
\begin{equation}
\mathcal{L}=\mathcal{L}_\mathrm{recon}+\lambda_\mathrm{lpips} \mathcal{L}_\mathrm{lpips}+\lambda_\mathrm{ssim} \mathcal{L}_\mathrm{ssim} + \lambda_\mathrm{dist} \mathcal{L}_\mathrm{dist}
\end{equation}

where $\lambda_\mathrm{lpips}$, $\lambda_\mathrm{ssim}$ and $\lambda_\mathrm{dist}$ are the scale to balance the perceptual, structural similarity and distortion regularization respectively. In our experiments, they are set to be 0.1, 0.1 and 0.01.

\section{Experiments}
\label{sec:exp}

\paragraph{Training Details}
\label{sec: training details}
We employ the Adam optimizer with a learning rate of 0.0001, $\beta_1$ = 0.9, and $\beta_2$ = 0.999. To ensure stable training in the early stages, we also implement a learning rate warm-up. Our model is trained on 32 NVIDIA A100 GPUs with a batch size of 128 for 1000 epochs at $128 \times 72$ and $512 \times 288$ resolutions, respectively. In addition, for each input sequence, we define the triplane orientation using the camera direction of the middle frame in the source sequence.

\paragraph{Datasets}
We conduct experiments on several datasets:

$\underline{\textit{NVIDIA Dynamic Scenes}}$~\citep{yoon2020novel}: The NVIDIA Dynamic Scenes dataset is a widely used benchmark for evaluating dynamic scene synthesis. It comprises eight dynamic scenes captured by a synchronized rig with 12 forward-facing cameras. We derive monocular videos following the setup in prior works~\citep{li2023dynibar, zhao2024pseudo}, ensuring that the resulting monocular videos cover most timesteps. 

$\underline{\textit{EPIC Fields}}$~\citep{tschernezki2024epic}: The EPIC-KITCHENS is a comprehensive egocentric dataset. EPIC Fields extends EPIC-KITCHENS by including 3D camera poses. This augmentation reconstructs 96\% of the videos in EPIC-KITCHENS, encompassing 19 million frames recorded over 99 hours in 45 kitchens. To minimize redundancy and skew while ensuring sufficient viewpoint coverage, we apply the frame filtering method from \cite{tschernezki2024epic} to extract monocular videos.

$\underline{\textit{nuScenes}}$~\citep{vora2020pointpainting}: The nuScenes dataset is a large-scale autonomous driving benchmark, comprising 1000 scenes, 
pre-split into training and test sets. Each sample includes RGB images from six cameras, providing a 360° horizontal field of view. For our experiments, we use only the three forward-facing camera views to extract monocular videos. We adopt the dataset split from \cite{vora2020pointpainting} for generalization testing.

$\underline{\textit{Plenoptic Video dataset}}$~\citep{li2022neural}: The Plenoptic Video dataset is a real-world dataset captured using a multi-view camera system consisting of 21 GoPro cameras operating at 30 FPS. Each scene in the dataset comprises 19 synchronized videos, each 10s in duration.

$\underline{\textit{DAVIS}}$~\citep{pera2016davis}:
The DAVIS dataset is a high-quality video object segmentation benchmark consisting of 50 video sequences with 24 FPS and Full HD resolution. We process video sequences  using the same settings as prior work \cite{Zhao2024PGDVS} and evaluate generalization ability on this dataset.

\paragraph{Evaluation}
Our prediction yields a future 3D representation for which we do not have corresponding ground-truth (GT) data. Consequently, we can not directly assess the quality of 3D representation. To tackle this issue, we indirectly evaluate the quality of novel views rendered from the future 3D representation. We utilize Learned Perceptual Image Patch Similarity (LPIPS)~\citep{zhang2018unreasonable}, Structural Similarity Index (SSIM)~\citep{wang2004image}, and Peak Signal-to-Noise Ratio (PSNR) as the primary metrics for evaluation.

\paragraph{Baselines}
\label{baselines}
To our knowledge, no prior work has specifically addressed our task, making direct comparisons with existing methods challenging. The most relevant to our work is 4D reconstruction, however, existing methods in 4D reconstruction render novel views at target time using all video frames, not just the preceding ones. To align with the 3D future prediction setting, we adapt these 4D reconstruction methods by using previous video frames along with the video frame at the target time as input. In contrast, our approach restricts itself to using only previous frames. Specifically, we take PGDVS$^{\dagger}$ and GNT~\citep{wang2022attention} as baselines. PGDVS$^{\dagger}$  is a generalized variant of PGDVS ~\citep{Zhao2024PGDVS} that incorporates input depth from ZoeDepth.  GNT is a generalized reconstruction method and can take only monocular video frames as input. While GNT is primarily intended for static scenes, it can be applied to dynamic scenes by treating the input video frames as different views of “static” scenes, allowing it to output novel view of “static” scenes at the target time.

\begin{table}[t]\small
\centering
\caption{\textbf{Results on NVIDIA Dynamic Scenes.} $^{\dagger}$ denotes the generalized variant of PGDVS that incorporates input depth from ZoeDepth~\citep{bhat2023zoedepth}. LPIPS is reported by multiplying with 100.}
\setlength{\tabcolsep}{3.0pt}
\resizebox{\columnwidth}{!}{%
\begin{tabular}{lccccccccc}
\toprule
 \multirow{2}[1]{*}{Model}& \multicolumn{3}{c}{Full Image} & \multicolumn{3}{c}{Dynamic Area} & \multicolumn{3}{c}{Static Area} \\
\cmidrule(lr){2-10}
 &PSNR↑ & SSIM↑ & LPIPS↓ & PSNR↑ & SSIM↑ & LPIPS↓ & PSNR↑ & SSIM↑ & LPIPS↓ \\
\midrule
GNT &20.33 & 0.650 & 26.38 & 17.30 & 0.417 & 54.23 & 21.57 & 0.680 & 24.20 \\
PGDVS$^{\dagger}$ &19.23 & 0.630 & 25.84 & 15.89 & 0.334 & 52.26 & 20.38 & 0.665 & 21.70 \\
Ours &\textbf{22.43} & \textbf{0.706} & \textbf{16.29} & \textbf{18.64} & \textbf{0.652} & \textbf{33.04} & \textbf{24.03} & \textbf{0.724} & \textbf{15.79} \\
\bottomrule
\end{tabular}
\label{tab:nvs on nvidia}
}
\end{table}

\begin{table}[t]\small
\centering
\caption{\textbf{Results on DAVIS dataset.} LPIPS is reported by multiplying with 100.}
\setlength{\tabcolsep}{3.0pt}
\resizebox{\columnwidth}{!}{%
\begin{tabular}{lcccccccccc}
\toprule
 \multirow{2}[1]{*}{Model}  & \multicolumn{3}{c}{Full Image} & \multicolumn{3}{c}{Dynamic Area} & \multicolumn{3}{c}{Static Area} \\
\cmidrule(lr){2-10}
 &PSNR↑ & SSIM↑ & LPIPS↓ & PSNR↑ & SSIM↑ & LPIPS↓ & PSNR↑ & SSIM↑ & LPIPS↓ \\
\midrule
GNT &19.10 & \textbf{0.547} & 39.99 & 14.44 & 0.306 & 72.50 & 20.12 & \textbf{0.567} & 35.28 \\
Ours &\textbf{21.05} & 0.535 & \textbf{32.42} & \textbf{17.81} & \textbf{0.441} & \textbf{53.34} & \textbf{21.50} & 0.537 & \textbf{31.00} \\
\bottomrule
\end{tabular}
\label{tab:nvs on davis}
}
\end{table}

\begin{figure*}[t]
	\begin{center}
		\includegraphics[width=1.\linewidth]{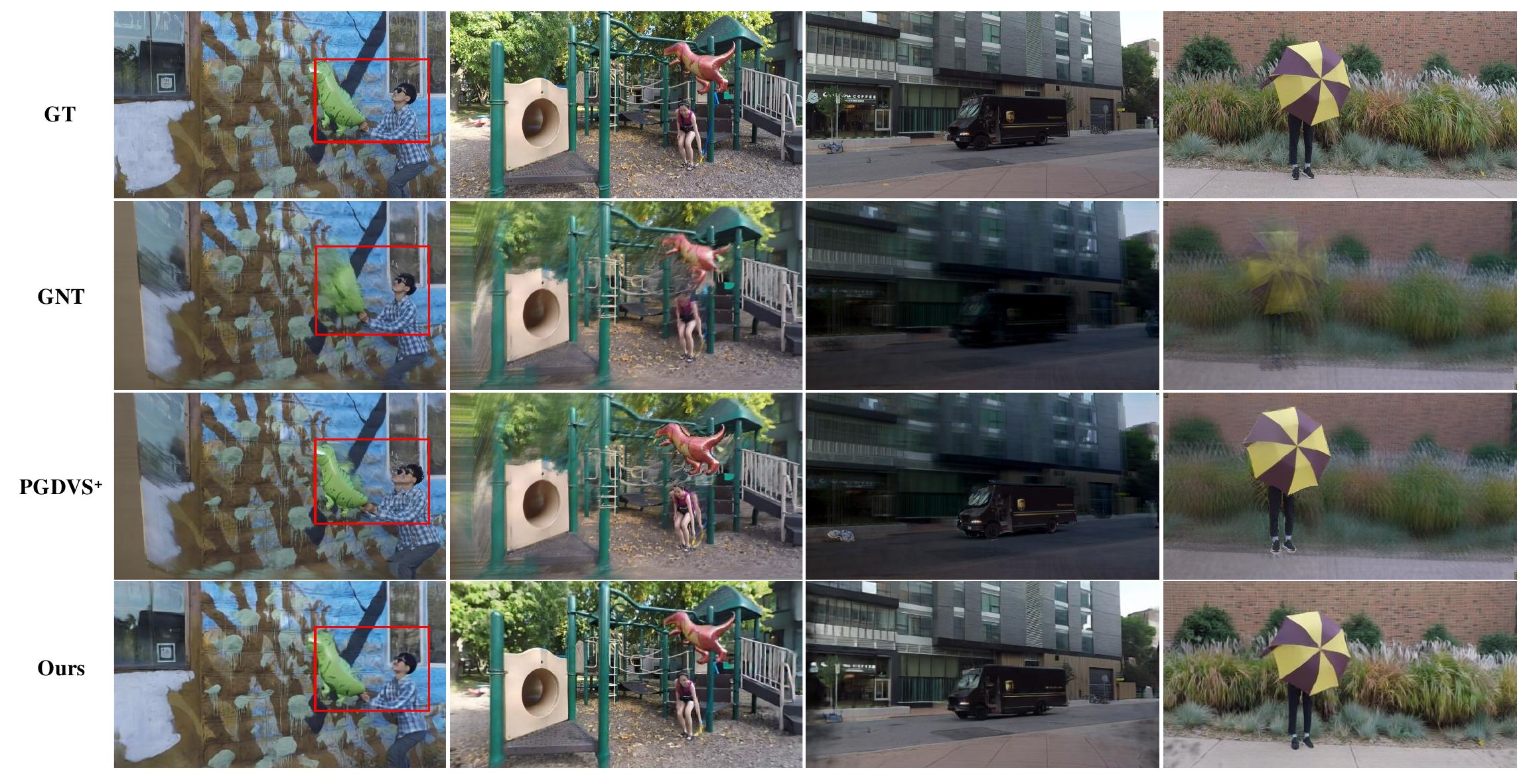}
	\end{center}
	\caption{\textbf{Qualitative Comparison on the NVIDIA Dynamic Scenes Dataset.} Our method significantly outperforms GNT and PGDVS$^{\dagger}$ in both dynamic and static content. For dynamic objects (first two columns), our approach delivers precise motion and avoids motion blur compared with PGDVS$^{\dagger}$. For static contents (last two columns), our method shows a clear background, whereas PGDVS$^{\dagger}$ and GNT result in blurred backgrounds due to limited depth priors.}
	\label{fig:compare}
	\vspace{-4mm}
\end{figure*}

\subsection{Main Results}

In this section, we present both quantitative(\Cref{sec:quantitative}) and qualitative results(\Cref{sec:qualitative}) to demonstrate the effectiveness of our proposed method. We utilize five datasets: NVIDIA, EPIC, Plenoptic, nuScenes, and DAVIS.  Our model is trained on scenes from EPIC, Plenoptic, and the nuScenes train set, and is subsequently evaluated on the other two datasets, NVIDIA and DAVIS, as detailed below.

\subsubsection{Quantitative Evaluation}
\label{sec:quantitative}
\paragraph{Setup}
We conduct quantitative evaluations on the NVIDIA Dynamic Scenes and DAVIS datasets. For our task, which involves next-time prediction, all next-time views can be treated as novel views of the same scenes. On the NVIDIA dataset, each scene frame consists of 12 views and all views are used for testing. For the DAVIS datasets, we use COLMAP to estimate camera poses and evaluate novel view synthesis on a single view. As mentioned in~\cref{baselines}, PGDVS$^{\dagger}$ needs one view at target time as input, thus this baseline can not be applied to the DAVIS dataset for comparison. 

\paragraph{Results}
On the NVIDIA Dynamic Scenes dataset, we compare our method to GNT~\citep{wang2022attention} and PGDVS~\citep{Zhao2024PGDVS}. As shown in~\Cref{tab:nvs on nvidia}, our method outperforms both GNT and PGDVS$^{\dagger}$ across all metrics. Specifically, our method achieves a superior PSNR of 18.64 for the dynamic components of the scenes, compared to GNT (17.30) and PGDVS$^{\dagger}$ (15.89). This highlights that the strong 3D future prediction of our approach. We also observe that PGDVS$^{\dagger}$ performs worse on the PSNR metric than GNT, despite being based on it. We conjecture this may be due to the shifting of dynamic content, such as persons and cars, see~\Cref{fig:compare}. For the static components, our method significantly surpasses both PGDVS$^{\dagger}$ and GNT on the PSNR and LPIPS metrics. The poor performance of GNT and PGDVS$^{\dagger}$ can be attributed to the depth scale-shift ambiguities~\citep{Zhao2024PGDVS}. In contrast, our high performance indicates that our method is robust against depth scale-shift ambiguities. 

For DAVIS dataset, our method consistently beats GNT across almost all metrics. As shown in~\Cref{tab:nvs on davis}, our method achieves a PSNR of 17.81 for the dynamic components of the scenes, significantly exceeding GNT. For static areas, our method also surpasses GNT on both PSNR and LPIPS metrics, although it yields slightly worse results on the SSIM metric. We speculate that this may be caused by the joint modeling of our method for dynamic and static areas. Unlike GNT, the performance of our method on static content is influenced by dynamic content. In addition, it is important to note that there is a significant domain gap between our training dataset and the DAVIS dataset. The high performance on the DAVIS dataset demonstrates that our method exhibits strong generalization across diverse scenarios.

\begin{figure*}[t]
	\begin{center}
		\includegraphics[width=1.\linewidth]{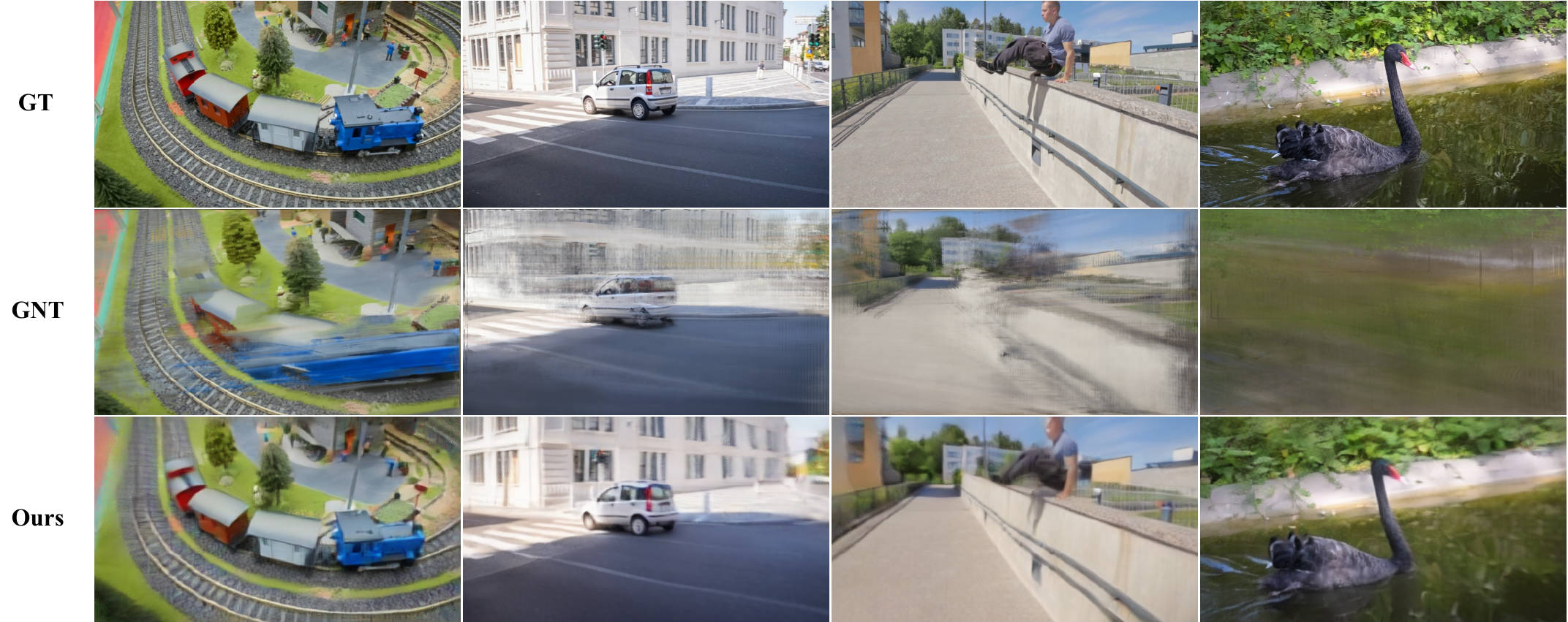}
	\end{center}
	\caption{\textbf{Qualitative Comparison on the DAVIS Dataset}. Our model produces high-quality novel views on DAVIS dataset, including indoor, outdoor, dynamic, and static settings. Notably, the blackswan (last column) is not present in our training data, showing the robust generalization capabilities of our method.}
	\label{fig:nvs}
	\vspace{-4mm}
\end{figure*}

\subsubsection{Qualitative Evaluation}
\label{sec:qualitative}

We conduct a qualitative evaluation using the NVIDIA Dynamic Scenes and DAVIS dataset. As illustrated in~\Cref{fig:compare}, our method outperforms GNT and PGDVS$^{\dagger}$ in both dynamic and static content on NVIDIA Dynamic Scenes. For dynamic content, our method precisely captures object motion without omitting details, such as the person's head and legs (column 1). Additionally, our model effectively avoids motion blur in dynamic objects, like the person body (column 4). For static content, our model excels in background modeling without relying on depth priors. PGDVS$^{\dagger}$ performs worse due to its dependence on limited depth priors, highlighting the drawbacks of relying on pre-trained models. 

\Cref{fig:nvs} demonstrates the generalization capability of our framework on the DAVIS dataset. Although our model is trained only with kitchen and driving scenarios, it performs well on out-of-distribution scenes. In column 4, our model accurately predicts the motion of a blackswan, showcasing its acquisition of general 4D prior knowledge. For scenarios (column 1 and 2) similar to our training data, our model exhibits superior performance. These results suggest that our model could achieve even better outcomes with more data. 

\subsection{Ablation Study}

\begin{table}[t]\small
\centering
\caption{\textbf{Ablation studies on NVIDIA Dynamic Scenes at $128 \times 72$ resolution.} We analyze the effects of various components, focusing on the influence of different model architectures and different loss functions on the overall performance. LPIPS is reported by multiplying with 100.}
\setlength{\tabcolsep}{0.8pt}
\begin{tabular}{cccccccccc}
\toprule
 \multirow{2}[1]{*}{Model}& \multicolumn{3}{c}{Full Image} & \multicolumn{3}{c}{Dynamic Area} & \multicolumn{3}{c}{Static Area} \\
\cmidrule{2-10}
 & PSNR↑ & SSIM↑ & LPIPS↓ & PSNR↑ & SSIM↑ & LPIPS↓ & PSNR↑ & SSIM↑ & LPIPS↓ \\
\midrule
w/o Temporal-based 3D Constraint & 25.78 & 0.830 & 5.60 & 20.71 & 0.691 & 35.1 & 27.34 & 0.857 & 4.84 \\
w/o Self-attention & 27.68 & 0.854 & 5.12 & 21.98 & 0.695 & 34.2 & 28.64 & 0.867 & 4.27 \\
w/o Plane-attention & 27.93 & 0.871 & 4.54 & 23.76 & 0.797 & 16.9 & 29.01 & 0.883 & 2.78 \\
\midrule
w/o LPIPS Loss & 29.80  & 0.914 & 5.59 & 25.13 & 0.857  & 16.3 & 30.97 & 0.922 & 2.68     \\
w/o SSIM Loss     & 27.51 & 0.857 & 4.17 & 23.09 & 0.765 & 18.5 & 28.73 & 0.872 & 2.71      \\
w/o Distortion Loss     & 28.43 & 0.888 & 3.72 & 24.12 & 0.822 & 14.9 & 29.58 & 0.899 & 2.30  \\
\midrule
Ours& 28.56 & 0.884 & 4.25 & 24.27 & 0.810 & 15.0 & 29.78 & 0.896 & 2.47 \\
\bottomrule
\end{tabular}
\label{tab:ablation}
\vspace{-4mm}
\end{table}

In this section, we conduct ablation studies on the NVIDIA Dynamic Scenes dataset to better understand the contributions of different components to the efficacy of our approach.
Due to computational resource limitations, these studies utilizes images with a resolution of $128 \times 72$, employing a batch size of 32 for 500 epochs during training. Qualitative results are provided in~\Cref{sec: ablation results} for further reference.
Additionally, we detail the static/dynamic detection mechanism of the temporal-aware view-attention in~\Cref{sec: view-attention}.

\paragraph{Temporal-based 3D Constraint}
To assess the impact of our temporal-based 3D constraint, we establish a baseline that applies only photometric loss on the target view frame at time $t=S+1$ during training. The experimental results in~\Cref{tab:ablation} demonstrate that the temporal-based 3D constraint significantly improves the performance across all metrics. This strategy leverages the disparity between two target views to impose geometric constraints on the generated triplanes, leading to more accurate 3D consistency. In contrast, the approach based on single target view lacks this constraint and suffers from scale ambiguity, resulting in noticeable pixel shifts in the rendered images.

\paragraph{Self-attention in Image Encoder}
We explore the impact of the self-attention module by removing it from the image encoder. The results, detailed in~\Cref{tab:ablation}, show a significant decrease in novel view synthesis metrics, especially in the synthesis of dynamic objects. This decrease stems from the lack of long-range context in the image encoder. The self-attention module addresses this issue and enhances motion-aware feature aggregation, highlighting its critical role in the dynamic scene synthesis.

\paragraph{Plane-attention}
To evaluate the effect of plane-attention, we conduct an ablation study on this component. The results in~\Cref{tab:ablation} indicate that plane-attention can improve our model across all metrics, validating its benefit in improving triplane features.

\paragraph{Impact of Losses}
As shown in~\Cref{tab:ablation}, the absence of LPIPS loss results in a noticeable drop on the LPIPS metric, demonstrating its effectiveness. Nevertheless, our model retains favorable LPIPS values even without LPIPS, underscoring its inherent capability to capture perceptual quality. The SSIM loss boosts performances on PSNR and SSIM, aiding in the learning of low-level high-frequency details. Although distortion loss negatively affects SSIM and LPIPS, it addresses the "floater" issue in rendered image, resulting in a positive gain on PSNR (Full Image).

\begin{figure*}[t]

	\begin{center}
		\includegraphics[width=1.0\linewidth]{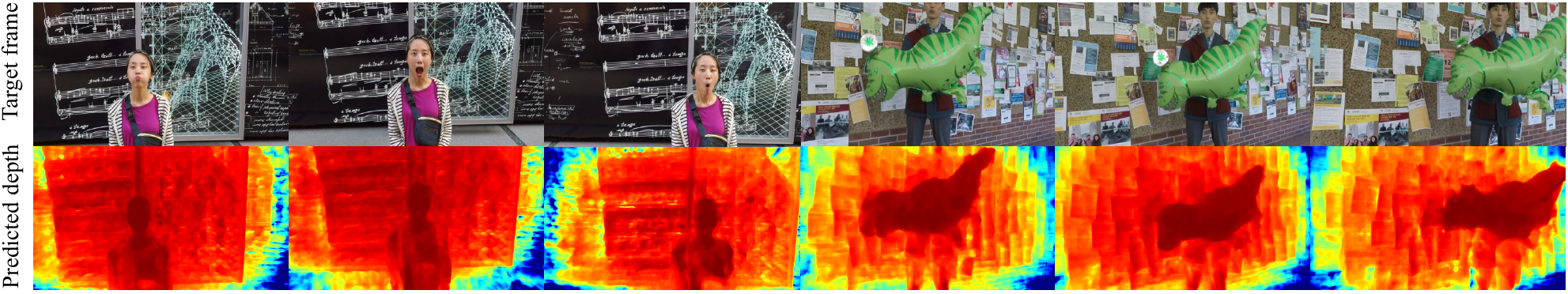}
		\vspace{-5mm}
	\end{center}
    \setlength{\abovecaptionskip}{0.mm}
	\caption{\textbf{Reconstructed depth maps on NVIDIA Dynamic Scenes.} Red indicates closer distances, while blue denotes farther distances.}
	\label{fig:depth}
\end{figure*}

\begin{figure*}[t]
    \begin{center}
		\includegraphics[width=1.\linewidth]{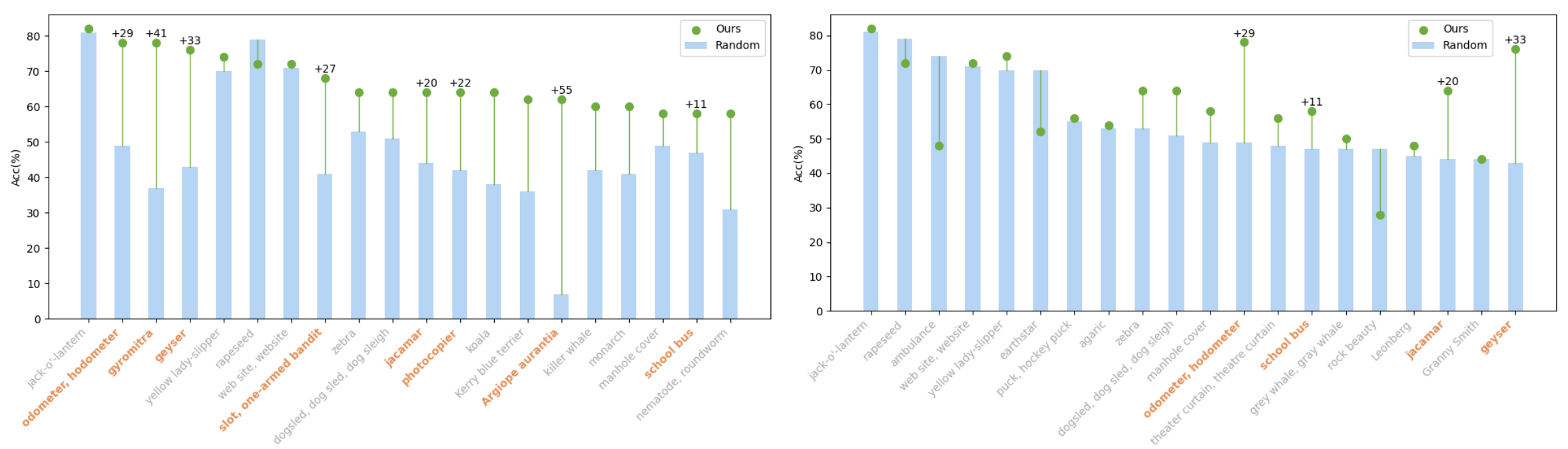}
		\vspace{2mm}
	\end{center}
    \setlength{\abovecaptionskip}{-8.mm}
	\caption{\textbf{Comparisons on ImageNet linear classification}: our model vs a random-initialized model. The highlighted categories (\textcolor{orange}{orange}) are closely related to our training data. The categories in left and right parts are selected by top-1 classification accuracy of our model and a random-initialized model, respectively. We show the top 20 categories.}
	\label{fig:imagenet}
	\vspace{-2mm}
\end{figure*}

\subsection{Emergent Capabilities}

\paragraph{Geometry Learning}
We illustrate the depth maps of our model on NVIDIA Dynamic Scenes in~\Cref{fig:depth}. The results indicate that our model can predict depth through self-supervised learning, even without prior knowledge from pre-trained models.
Regarding the blocky appearance of the depth map on the right side of~\Cref{fig:depth}, we find these blocky areas are closely related to posters. We conjecture that our model can distinguish each individual poster but does not recognize that these posters are on the same wall plane.
Although the current depth results are not perfect, we expect this capability to improve as more data becomes available.
\paragraph{Semantic Learning}
\label{sec:sl}
To validate the representation learning of our method, we report the linear probing top-1 accuracy of our image encoder on ImageNet. We use a random-initialized model as the baseline. 
See detailed experimental settings in~\Cref{sec: imagent}.
As shown in \Cref{fig:imagenet}, the highlighted categories (\textcolor{orange}{orange}), such as odometer and school bus, are closely related to our training data. Notably, our model significantly outperforms the baseline on these categories, indicating effective learning of semantic information from the training data. This suggests that our method is a promising representation approach as the training data increases.

\section{Conclusions}
We present a method for predicting 3D representations at future times using only monocular video streams. This method is trained on large-scale monocular videos in a self-supervised manner. Our model demonstrates strong generalization on unseen scenarios. Furthermore, the emergent capabilities indicate that our approach could pave the way for advancements in spatial intelligence. We hope our work will inspire more future research into next-3D prediction.

\bibliography{ref}
\bibliographystyle{iclr2025_conference}

\newpage
\appendix
\section{Appendix}
\label{appendix}

\subsection{Architecture details}
\paragraph{Image Encoder}
The image encoder consists of a ResNet-like backbone and a self-attention layer. The ResNet component includes three downsampling layers and nine ResNet blocks. Additionally, the self-attention layer utilizes 2D sinusoidal position encoding, with a channel dimension of 576.

\paragraph{Camera Encoder}
The camera encoder is composed of a straightforward linear layer. Camera intrinsics are processed using a sinusoidal encoding function before being fed into the camera encoder. This linear mapping aligns the channel dimensions of the camera features with those of the image features, facilitating consistent feature integration.

\paragraph{4D-aware Transformer}
Feature aggregation is accomplished through a stack of 12 basic 4D-aware Transformer modules, each with an output dimension of 576.

\paragraph{Upsampler}
The upsample module consists of a single deconvolution layer that scales the triplane from $32 \times 32 \times 576$ to $128 \times 128 \times 192$.  Given the interdependent nature of the planes, we adopt an approach inspired by Rodin~\citep{wang2023rodin}. Specifically, for each pixel in a plane, its feature is concatenated with the average feature of the corresponding row or column from the other two planes, thereby enhancing contextual integration across the triplane structure.

\paragraph{Triplane Decoder}
The triplane decoder employs a simple two-layer MLP. Position encoding for the triplane features follows the methodology outlined in~\citep{wang2023pet}.

\begin{figure*}[t]

	\begin{center}
		\includegraphics[width=1.0\linewidth]{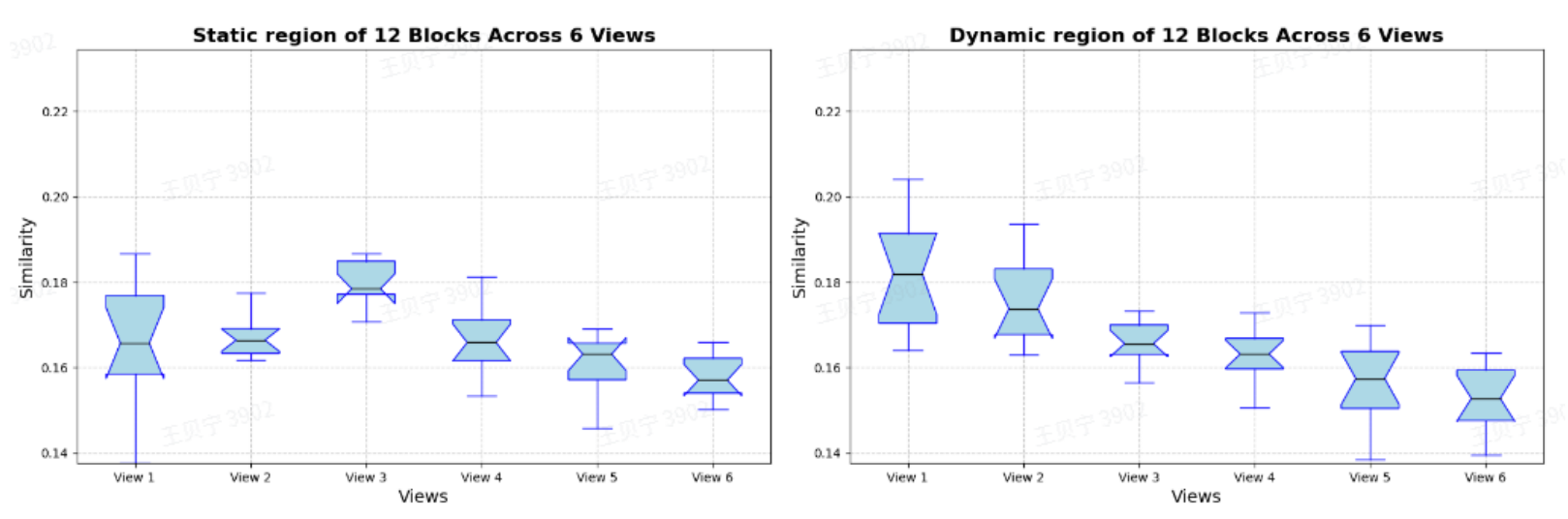}
		\vspace{-5mm}
	\end{center}
    \setlength{\abovecaptionskip}{0.mm}
	\caption{\textbf{Boxplot summarizing the similarity measures of dynamic and static 3D points.}The x-axis represents 6 views from the source frames, ordered by increasing temporal distance from the target view.}
	\label{fig:view-attention}
	\vspace{-3mm}
\end{figure*}
\subsection{The static/dynamic detect mechanism in Temporal-aware View-attention}
\label{sec: view-attention}
\paragraph{Setup}
We explain the static/dynamic detect mechanism of temporal-aware view-attention through an analysis of a target image from the NVIDIA "playground" scene.
First, pixels from the target image are transformed into 3D points based on predicted depths and classified into static/dynamic points using the semantic mask.  Next, we compute the similarities between these 3D points and their corresponding epipolar points from six source views/times across 12 temporal-aware view-attention blocks. Finally, we calculate the mean and variance of similarities for static and dynamic points, respectively.

\paragraph{Results}

As shown in~\Cref{fig:view-attention}, dynamic 3D points exhibit high similarity at close time and low similarity at more distant time. In contrast, static points maintain a consistent level of similarity across different times/views. This phenomenon indicates that the temporal-aware view-attention effectively distinguishes between dynamic and static points based on similarity.

\subsection{Qualitative comparison of ablation studies on NVIDIA scenes}
\label{sec: ablation results}
\begin{figure*}[h]
	\begin{center}
		\includegraphics[width=1.\linewidth]{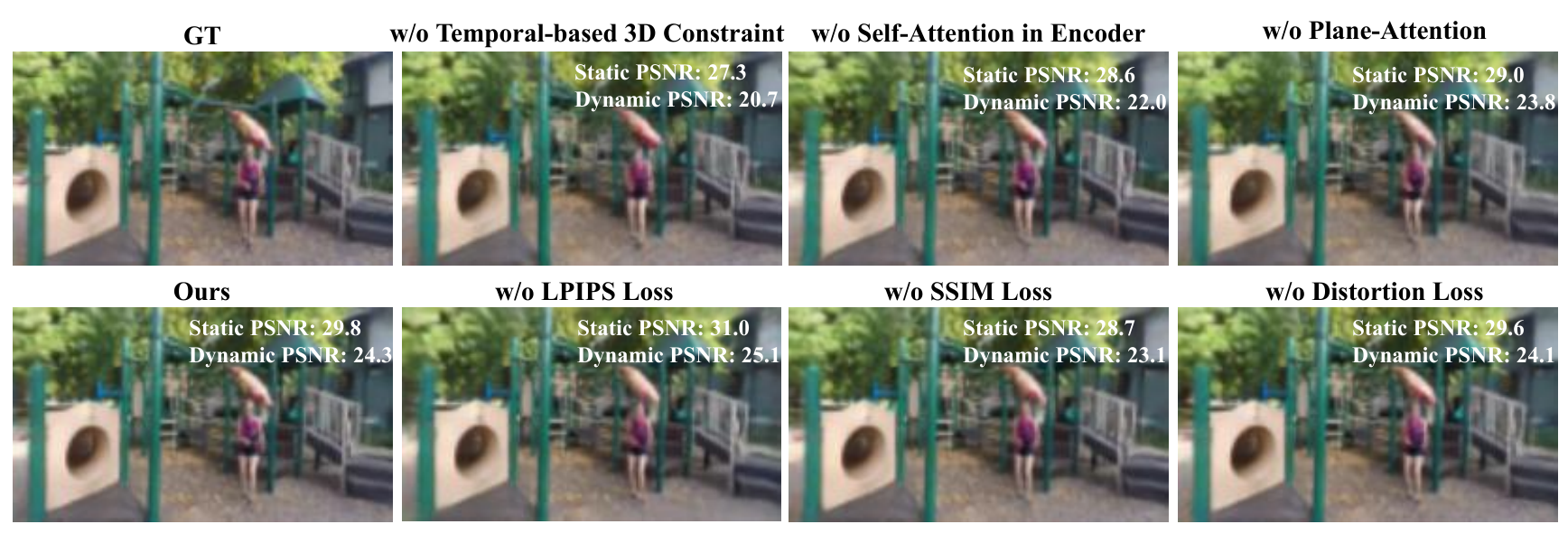}
		\vspace{-4mm}
	\end{center}
    \setlength{\abovecaptionskip}{0.mm}
	\caption{\textbf{Qualitative comparison of ablation studies on  NVIDIA Dynamic Scenes dataset at $128 \times 72$ resolution.} The metrics mean the PSNR on the testing dynamic and static contents in different ablation studies.}
	\label{fig:ablation}
	\vspace{-4mm}
\end{figure*}

We evaluate the impact of temporal-based 3D constraint, self-attention in the image encoder, plane-attention, and the applied loss functions, with the qualitative results presented in~\Cref{fig:ablation}.
\subsection{Comparisons on ImageNet linear classification}
\label{sec: imagent}
To evaluate the semantic learning capabilities of our model, we conduct a linear classification experiment on ImageNet using our image encoder. It is important to note that the encoder is trained without LPIPS loss to prevent semantic leakage. We set a randomly initialized image encoder as the baseline. To ensure the stability of the experimental results, each model is trained for 3 times with different seeds and the averaged top-1 classification accuracy is used as the final result.

\subsection{Limitations}
\label{sec: limitations}
Our method has three limitations. First, it requires camera intrinsics and extrinsics for training. These camera parameters may not align well with the ground truth. As an alternative, we can consider optimizing both the camera parameters and our model jointly in a data-driven manner. Second, our model is deterministic, which limits its performance on unseen content that is not present in the source views. To address this, we plan to introduce diffusion models to generate unseen content in the future. Lastly, due to limited resource, we are unable to train a large model with large-scale datasets (\eg Ego4D~\citep{grauman2022ego4d}) to validate the scalability of our approach.

\end{document}